\documentclass[runningheads]{llncs}
\usepackage{amsmath}
\usepackage{graphicx}
\begin{document}
\title{Rejection-Cascade of Gaussians: Real-time adaptive background subtraction framework}
\titlerunning{Rejection-Cascade of Gaussians}
\author{B Ravi Kiran$^1$ \and
Arindam Das$^2$ \and
Senthil Yogamani$^3$}
\authorrunning{Kiran et al.}
%
\institute{
Navya, Paris, France \\ \and
Detection Vision Systems, Valeo India \\ \and
Valeo Vision Systems, Galway, Ireland \\
\email{ravi.kiran@navya.tech, \{arindam.das,senthil.yogamani\}@valeo.com}}

\maketitle              

\begin{abstract}
Background-Foreground classification is a well-studied problem in computer vision. Due to the pixel-wise nature of modeling and processing in the algorithm, it is usually difficult to satisfy real-time constraints. There is a trade-off between the speed (because of model complexity) and accuracy. Inspired by the rejection cascade of Viola-Jones classifier, we decompose the Gaussian Mixture Model (GMM) into an adaptive cascade of Gaussians(CoG). We achieve a good improvement in speed without compromising the accuracy with respect to the baseline GMM model. We demonstrate a speed-up factor of 4-5x and 17 percent average improvement in accuracy over Wallflowers surveillance datasets. The CoG is then demonstrated to over the latent space representation of images of a convolutional variational autoencoder(VAE). We provide initial results over CDW-2014 dataset, which could speed up background subtraction for deep architectures.
\end{abstract}

\section{Introduction}
\label{sec:intro}
Background subtraction is critical component of surveillance applications (indoor and outdoor), action recognition, human computer interactions, tracking, experimental chemical procedures that require significant change detection. Work on background subtraction started since the 1970s and even today it is an active open problem. There have been a host of methods which have been developed and below is a short review which will serve to aid understanding our algorithm. A survey by ~\cite{Piccardi} provides an overview of common methods which includes Frame differencing (FD), Running Gaussian average (RGA), Gaussian Mixture Model (GMM) and Kernel Density Estimation (KDE). We employ these basic methods in a structured methodology to develop our algorithm.

A survey of variants of GMM, issues and analysis are presented in ~\cite{Bouwmans}. In our work, we focus on solving the variable-rate adaptation problem and improving the performance. Abstractly, our work tries to fuse several algorithms to achieve speed and accuracy and we list similar methods here. Similar attempts have been made by the following researchers. ~\cite{toyama} and ~\cite{Mubarak} used a Hierarchical background subtraction method that operates in different scales over the image : namely pixel, region and image level, while their models themselves are not hierarchical. Authors ~\cite{Zuo} switch between GMM and RGA models, while choosing a complex model for complicated backgrounds and simple model for simpler backgrounds. They use an entropy based measure to switch between the different models.
We briefly describe our observations and improvement over the standard GMM from ~\cite{Strauffer}. We observe in most cases, background subtraction is an asymmetric classification problem with probability of foreground pixel being much lesser than that of background. This assumption fails in the case of scenes like highways, a busy street, etc. In our work, we focus mainly on surveillance scenarios where there is very low foreground occupancy. Our framework exploits this fact and at the same time handles variable rate changes in background and improves accuracy. Our key contributions in this paper include:
1. Decomposition of GMM to form an adaptive cascade of classifiers - Cascade of Gaussians (CoG) which handles complex scenes in an efficient way to obtain real-time performance. 2. A confidence estimate for each pixel’s classification which would be used to vary the learning rate and thresholds for the classifiers and adaptive sampling. 3. Learning a time windowed KDE from the training data-set which would act as a prior to the Adaptive Rejection Cascade and also help the confidence estimate.

The decomposition of the GMM into the cascade is similar to the increasing true positive detection rate inspired by the Viola Jones Rejection Cascade ~\cite{Viola}. Authors ~\cite{Valentine} provided an optimized lookup for highly probable colors in the incoming background pixels thus providing speedup in the access. 

\section{Components of the Cascade}
This section describes the different components of the rejection cascade and how they were determined. The rejection cascade is accompanied by the confidence measure to make an accurate background classification at each level of the cascade.

\textbf{Scene Prior in Background Model}: 
The process of distinguishing linearly varying background and noisy pixels is a challenge and critical since the background subtraction model intrinsically has no additional attribute to separate them. For this scenario, in our approach we introduce a prior probability for every pixel (eqn \ref{eqn:2}). The non-parametric probability distribution for the pixels assuming independent R,G,B channels is now given too. The Scene prior basically provides an non-parametric estimate of pixel-values value over $N$ frames during training. The choice of $N$ is empirical and depends on how much dynamic background and foreground is present in the training frames. To obtain complete variability we choose as large N as possible. Henceforth we refer to Scene Prior as the prior. In the training phase we estimate the underlying temporal distribution of pixels by calculating the kernel function that approximates the said distribution.  Our case primarily concentrates on long surveillance videos with sufficient information (minimal foreground) available in the training sequence that decides $N$. For the standard GMM model(assuming the covariance matrix is diagonal) the updates of the parameters include:
\vspace{-0.5cm}
\begin{equation}
\begin{split}
P(I_n(x,y))  = & \sum_{i=1}^K \omega_{i,n}*\eta(I_n(x,y),\mu_{i,n},\sigma_{i,n},\\ 
\omega_{n+1,k}(x,y)   \longleftarrow & (1-\alpha)\omega_{n,k}(x,y) + \alpha(M_{i,n+1})\\
\mu_{n+1,k}(x,y)   \longleftarrow & (1-\rho)\mu_{n,k}(x,y) + \rho I_n(x,y)\\
\sigma_{n+1,k}^2(x,y)   \longleftarrow & (1-\rho)\sigma_{n,k}^2(x,y) + \rho I_n(x,y) 
\end{split}
\label{eqn:2}
\end{equation}

Where $K_{\sigma}$, represents the gaussian kernel and $\sigma$ the scale or bandwidth. This Kernel function is calculated to provide the modes of the different pixels. Where $\eta$ represents the pixel mode distribution obtained in equation \ref{eqn:2}, where $\omega_i$ represents the ratio of the component i in the distribution of pixel $I_n(x,y)$, and $\mu_i$,$\sigma_i$  are the parameters of the component, $M$ represents 0 or 1 based on a component match and finally $\alpha$ represents the learning rate of the pixel model. The $\alpha$ is intialised for all pixels usually, there has been work in adapting it based on the pixel entropy. We use the pixel gradient value distribution to do the same. 

\textbf{Determining Learning Rate Hyper-parameters}:
 Besides the kernel density, we also estimate the dynamic nature of the pixels in the scene. This is obtained by the clustering the residue between consecutive frames into 3 categories : into static/drifting, oscillating and dynamic pixels (Fig \ref{fig:Cog} top right). This helps resolve a pixel drift versus a pixel jump as shown in example below in figure. Once we have the residue $R_n(x,y) = I_{n}(x,y) - I_{n-1}(x,y), n \in [1,N]$, we evaluate the normalized histogram over the residue values. We select bins intervals to extract the 3 classes based on the dynamic nature of pixels. A peaky first bin implies near zero residue, thus a drift or static pixels. A peaky second bin implies oscillating pixels and the other cases are considered as dynamic pixels. Based on these values we choose the weights for the confidence measure (explained in the next section). This frequency over each bin sets the learning rate for the pixel. The process of obtaining the right learning rates(confidence function) from the normalized binned histogram values to determine $\alpha, \beta$ and $\gamma$ test for the learning rates have determined empirically by shape matching the histograms. 
 
 \textbf{Clustering Similar Background - Spatio-Temporal Grouping}:
The next step in the training phase is to determine background regions of pixels, in the frame that behave similarly in terms of adapted variance, number of modes, and optimally use fewer parameters and lesser instructions to update this specific region's, pixel models. The problem definition can be formalized as: We are given $Nx(framesize)$ pixels and for each pixel $I_n(x,y)$ we have a set of matches of the form $(I_n(x,y), I_n(x',y'))_{t_n}$, which means that pixel $I_n(x,y)$ correlated with pixel $I_n(x',y')$ at frame number $n$ . From these N matches, we construct a discrete time series $x_i(t)$ by clustering pixel $F_x,y^n$ at time interval $t$ frames. A time series of the pixel $I_n(x,y)$ values at frame $n_0$. Intuitively, $x_i$ measures the correlation in behavior of pixels over time window $t$. For convenience we assume that time series $x_i$ have the same length. We group together pixel value time series so that similar behavior is captured by similarity of the time series $x_i(t)$. This way we can infer which pixels have a similar temporal pattern variances and modalities, and we can then consider the center of each cluster as the representative common pattern of the group. This helps us cluster similar behaving pixels together. This is can be seen a spectral clustering problem as described in ~\cite{Azran}. We try a simpler approach here first by clustering the adapted pixel variances(matrix V) and weights(matrix R) of first dominant mode of pixels within a mixture model.
\begin{enumerate}
  \item Get N frames \& estimate pixel-wise $\mu(t), \sigma(t), \omega(t)$
  \item Form matrix whose rows are adapted variance and ranked weight observations, while columns are variables $V$ and $R$, $V(t_k,i) = I(t_k) , k=1:N$
  \item Obtain covariance matrices $R_{cov} = Cov(R), V_{cov} = Cov(V)$
  \item Perform K-means clustering with K=3 (for temporal pixel residue due to dynamic, oscillating, or drifting BG).
  \item Threshold for pixels within $0.7-0.5 \sigma$
  \item Calculate the KDE of given cluster \& the joint occurrence distribution and associated weight $\omega_1$, $\mu_1$ and $\sigma_1$ 
\end{enumerate}
where $\mu_1$ is first dominant common cascade level at grouped pixels. This suffers from the setback that the variances chosen temporally do not correspond to mean values associated with the maximum eigen value as obtained in case of Spectral Clustering. So we have the pixel variance and adapted weight (dominant mode) covariance matrices $ R(x_i,y_i) = Cov(Var(I_n(x_i,y_i)))$ and $W(x_i,y_i) = Cov(Var(W_n(x_i,y_i)))$. A single gaussian is fit over thresholded covariance matrices (Adapted variance and first dominant mode weight).
\begin{equation}
    \begin{split}
         r_n = & \mu_\text{advar}-\sigma_\text{advar}<var(R_\text{cov})<\mu_\text{advar}+\sigma_\text{advar}\\
         w_n = & \mu_\text{adw}-\sigma_\text{adw}<var(W_{cov})<\mu_\text{adw}+\sigma_\text{adw}
    \end{split}
    \label{eq:3}
\end{equation}

The parameters $\mu_\text{advar}$, $\sigma_\text{advar}$ and $\mu_\text{adw}$,$\sigma_\text{adw}$ represent the mean and standard deviation of the cluster of pixel variances and adapted weights of the first dominant modes. The fundamental clustering algorithm requires Data set $R_cov$ and $V_cov$, number of clusters - quantization of the adapted weights or variances, Gram matrix ~\cite{Azran}. One critical point to note here is that, when we do not choose to employ spatio-temporal grouping, and reduce the number of parameters and consequent updates, we can use the Scene Prior covariance estimation to increase the accuracy of the foreground detection. This is very similar to the background subtraction based on Co-occurrence of Image Variations. 

\textbf{Confidence Measure :}
The confidence measure is a latent variable use to aid the Rejection Cascade to obtain a measure of fitness for the classification of a pixel based on various criteria. The Confidence $C_n (x,y) $ for a pixel $I_n (x,y)$ is given by $C_n (x,y) = \alpha P(x,y) + \beta (\Delta_n I(x,y) + \gamma M(I_n (x,y))$.

Here, $M()$ represents the difference between the current pixel value $I_n(x,y)$ and the parameters of the  model occurring at the top of the ordered Rejection cascade described below, while $\Delta_n I(x,y) = I_n (x,y)-I_{n-1} (x,y)$. As seen in the ordered tree, the first set of parameters would be the first dominant mode - $(\mu_1+\sigma_1,\mu_1-\sigma_1)$. This is carried out based on the level in which the pixel gets successfully classified. $P()$ represents the probability of occurrence of the pixel from the KDE. The values of $\alpha$ $\beta$ and $\gamma$ are determined by the normalized temporal residue distribution (explained above). The physical significance and implications of $\alpha$ $\beta$ and $\gamma$- $\alpha$ says how confident the region is and regions that are stable (for example from the segments from clustering adapted variances and weights of training phase pixel models) would have high $\alpha$ values. While the value of $\beta$ determines how fast the pixel would need to adapt to new incoming values and this would mean a lower effect of the prior distribution. The final parameter $\gamma$ determines the consistency of the pixel belonging to a model and this would change whenever the pixels behavior is much more dynamic (as opposed to a temporal residue weighting it).

\textbf{Confidence based temporal sampling}:
Applying multiple modes of background classifiers and observing the consistency in their model parameters (mean, variance, and connectivity) we predict the future values of these pixels. A threshold on confidence function value determined by using stable regions(using region growing) as a reference is used to select the pixels both spatially and temporally. The description of the confidence measure is given in more detail in section 2.3.  The pixels with low confidence reflect regions R over the frame with activity and thus a high probability of finding pixels whose label are in transition (FG-BG). Thus by thresholding the confidence function we sub-sample the incoming pixels spatio-temporally. This intuition is when pixel values arriving now are within the first dominant mode's $0.7\sigma$ region, and even more so within the CHP level for a large number of frames, the confidence value saturates. The Region $R(x_i,y_i) = C_n(x_i,y_i)>C_{ScencePrior}(x_i,y_i)$ is just a thresholded binary map of this confidence value. This is demonstrated in the analysis in section 3.
    
\textbf{Cascade of Gaussians CoG}
The proposed method can be viewed as a decomposition of the GMM in an adaptive framework so as to reduce complexity and improve accuracy using a strong prior to determine the scenarios under which said gains can be achieved. The prior is used to determine the modality of the pixels distribution and any new value is treated as a new mean with variance model. The Cascade can be seen to consist of K Gaussians which are ordered based on the successful classification of the pixel. During steady state the ordered cascade conforms to the Viola Jones Rejection Cascade with decreasing positive detection rates.

The cascade is first headed by a Consistent Hypothesis Propagation (CHP) classifier which basically repeats the labeling process on the current pixel if its value is equal to the previous value (previous frame). This CHP classifier is then followed by an ordered set of Gaussians $\omega_i.\eta(\mu_i,\sigma_i)$ including the spatio-temporally grouped parameters. The tree ordering is different for different pixel and the order is decided based on the prior distribution (KDE) of the pixel and the temporal consistency of the pixel in the different levels. When the pixel values do not belong to any of the dominant modes based on the prior, we have scenario where the beta weight and gamma weight only considered and alpha is rejected (Prior Nullified).

\begin{figure}[t]
\begin{center}
   \includegraphics[width=0.65\linewidth]{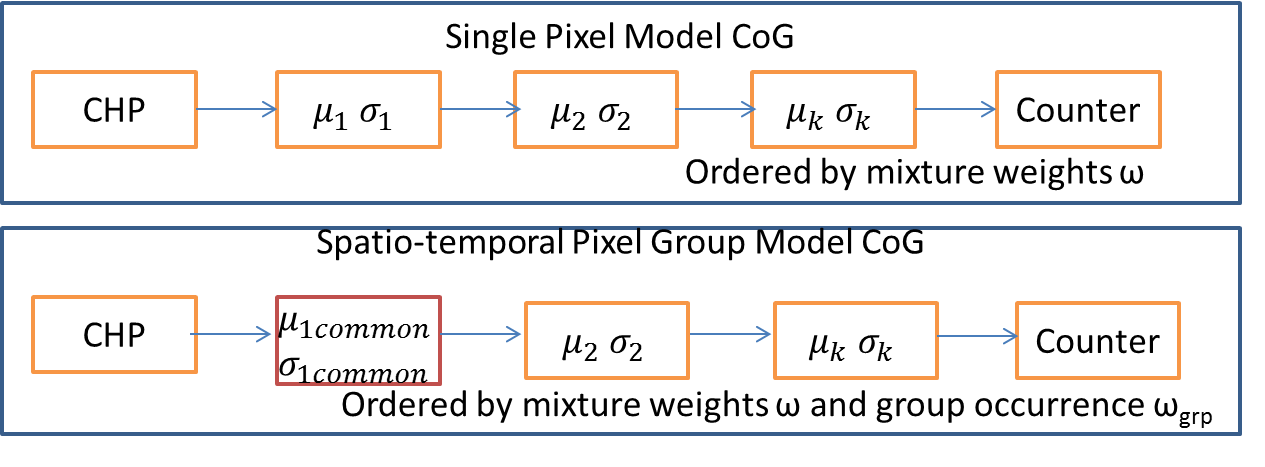}
   \includegraphics[width=0.3\linewidth]{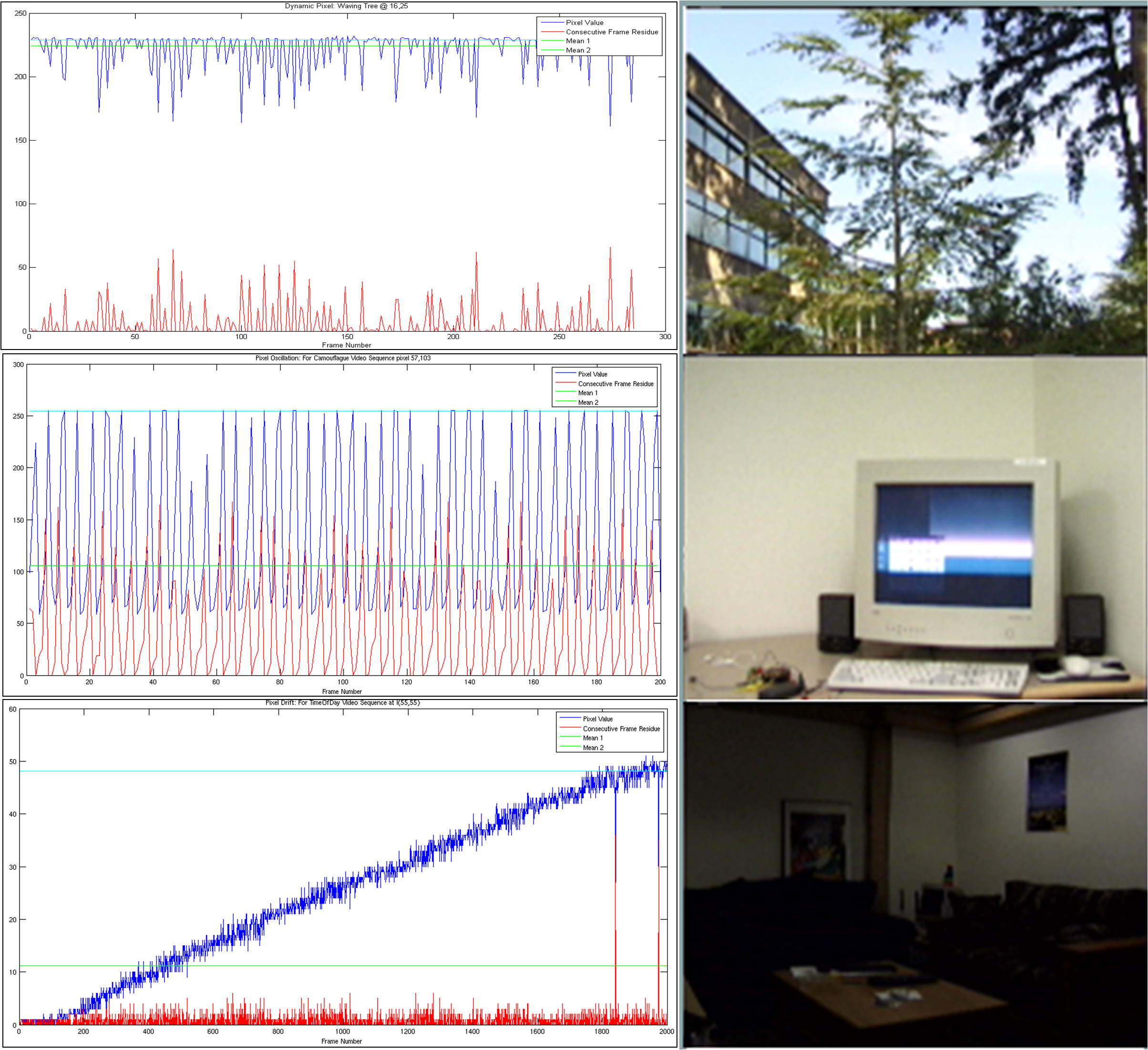}
   \includegraphics[width=.95\linewidth]{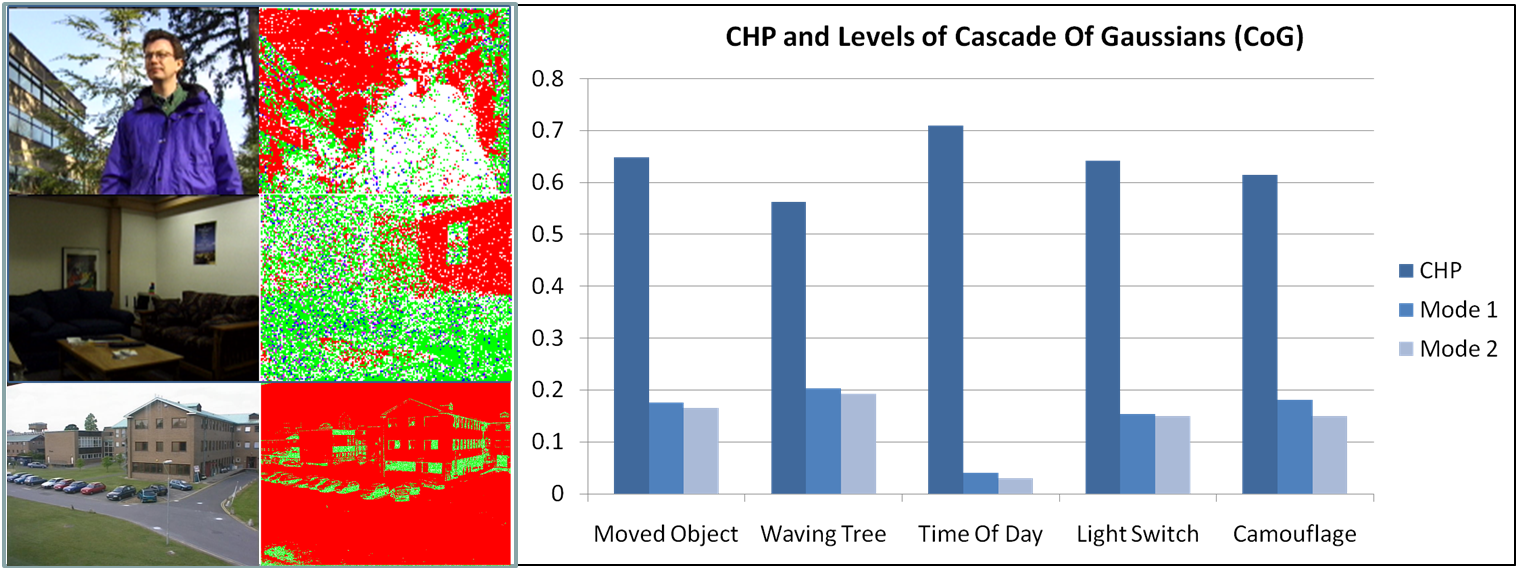}
   \caption{\textbf{Top-left} : Elements of CoG : CHP, first and second modes of gaussians and spatio-temporal window of CoG. \textbf{Top-right} : Dynamic Pixel Vs Oscillation Vs Pixel Drift, \textbf{Bottom} : 1. Pixels in CHP(red), Mode 1(green), Mode 2(blue),Mode 3(violet),Foreground(white) 2. Normalized pixel count over elements of Cascade of Gaussians CHP, first and Second modes of Gaussians. \vspace{-0.75cm}}
\end{center}

\label{fig:Cog}
\end{figure}

The rejection cascade assumes that the frequency of occurrences of foreground detections is lesser than that of the background. This idea was first introduced in the classic Viola-Jones paper ~\cite{Viola}. For the rejection cascade the training phase produces a sequence of features with decreasing rates of negative rejections. In our case we arrange the different classifiers in increasing complexity to maximize the speed. We observe in practice that, this cascade would also produce decreasing rates of negative rejections. The critical difference in this rejection cascade is that the classifier in each level of the cascade is evolving over time. To make adaptation efficient we adapt only the active level of the cascade, thus resulting in only one active update at a time, and during a transition the parameters are updated.

The performance of different rejection cascade elements is depicted in Figure \ref{fig:Cog}. It depicts cascade elements with increasing complexity (and consequently accuracy) have higher performance. These times were obtained over 4 videos from the wallflowers data set by ~\cite{toyama} of different types of dynamic background. This by itself can stand for the possible amount of speedup that can be obtained when the Rejection Cascade is operated on pixels adaptively based on the nature of the pixel.  In a similar observation we saw that the number of pixels (in each of these 4 videos) was distributed in different manner amongst the 4 levels. This is seen in figure \ref{fig:Cog}. Thus we see that even though the number of pixels corresponding to dynamic nature of pixel varies with the nature of the video, there is greater number of pixels on an average corresponding to low complexity Cascade elements.
The rejection cascade for BG subtraction was formed by determining (same as in ~\cite{Viola}) the set of background pixel classifiers (or in our case models like attentional operator in Viola Jones) and is organized as a degenerate tree such that it has decreasing false positive rate as we proceed down the cascade.  The performance of different rejection cascade elements are depicted in Figure \ref{fig:Cog}. It depicts cascade elements with increasing complexity (and consequently accuracy) have higher performance. These times were obtained over different types of static and dynamic background. This by itself can stand for the possible amount of speedup that can be obtained when the Rejection Cascade is operated on pixels adaptively based on the nature of the pixel.  In a similar observation we saw that the number of pixels (in each of these 4 videos) was distributed in different manner amongst the 4 levels. This is seen in figure \ref{fig:Cog}. Thus we see that even though the number of pixels corresponding to dynamic nature of pixel varies with the nature of the video, there is greater number of pixels on an average corresponding to low complexity Cascade elements. The learning rate for the model is calculated as a function of the confidence measure of the pixels. The abrupt illumination change is detected in the final level of the rejection cascade, by adding a conditional counter. This counter measures the number of pixels that are not modeled by the penultimate cascade element. If this value is above a threshold we can assume an abrupt illumination change scenario. This threshold is around seven tenth of the total number of pixels in the frame ~\cite{toyama}.

\section{Analysis \& VAE-COG}

\subsection{Scene Prior Analysis}
Here we discuss the the Scene Prior and its different components. First with regard to the clustering pixels based on their dynamic nature similarity, we show results of various clustering methods and their intuitions. The first model considers the time series of variances of said pixels in the N frames of training. The covariance matrix is calculated for the variances of the pixels. This can loosely act as the affinity matrix for the describing similar behavior of a pair of pixels. The weight of the first dominant mode is also considered to form the affinity matrix. 

\subsection{Cascade Analysis}
The CoG is faster on two accounts : Firstly it is cascade of simple-to-complex classifiers, CHP to RGA, and averaging over the performance (seen in figure), we see an improvement in speed of operation, since the simpler cases of classification outweigh the complex ones. Secondly it models the image as a spatio-temporal group of super pixels that needs a single set of parameters to update, even more so, when the confidence of the pixel saturates, the Cascade updates are halted, providing huge speedups. Though it is necessary to mention that the window of sampling is chosen empirically and in scale with the confidence saturation values. The average speedup of the rejection tree algorithm is calculated as : $\frac{I(x,y)}{\sum_i s_in_i}$ where x,y go over all indices of image, $n_i$ refers the ratio of background pixels labeled mean or mean with variance w.r.t the total number of background pixels in the image, $s_i$  is the normalized ratio of the time it takes for level $i$ BG model to evaluate and label a pixel as background. The values of $n$ and $s$ were profiled over various videos for different durations. Also we show the distribution of the CHP pixels as well as the first 3 dominant modes within different frames of Waving tree and Time of Day videos with 40 frames of training each. We can see a huge occupancy of Red (CHP) for both background and foreground pixels. Here we explain the confidence measure and effect on accuracy of the GMM model. We obtain a speedup of 2x-3x with the use of the Adaptive Rejection cascade based GMM. This speedup goes up at the effectiveness of accuracy of confidence based spatio-temporal sampling to 4-5x. This is evident in the Cascade level population (in figure \ref{fig:Cog}). We observe a 17\% improvement in accuracy over the baseline model because of adaptive modelling to handle difficult scenarios explicitly using scene priors. 

\begin{figure}[t]
\centering
\includegraphics[width=0.94\linewidth, angle=0]{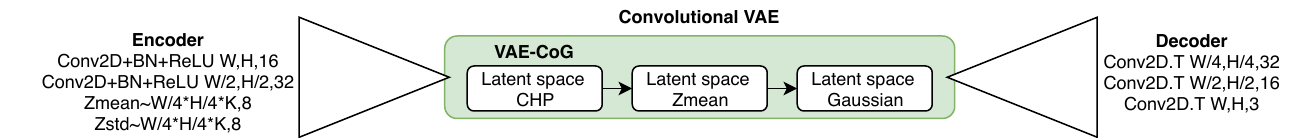}
\caption{The CoG rejection cascade over the latent space representation of the convolutional-VAE. The filters are all size 3x3.}
\label{fig:cogvae}
\end{figure}

\subsection{Latent space CoG with VAEs}
CNNs have become become the state-of-the-art models for various computer vision tasks. Our proposed framework is generic and can be extended to CNN models. In this section, we study a possible future extension of the the Rejection cascade to the Variational AutoEncoder
(VAE). There has been recent work on using auto-encoders to learn dynamic background for the subtraction task
\cite{Xu2014DyBGL}. Rejection cascades have also been employed within  convolutional neural networks architectures
for object detection \cite{yang2016exploit}. VAEs one of the most interpretable deep generative models.

\begin{figure}[t]
\begin{center}
\includegraphics[width=0.22\linewidth]{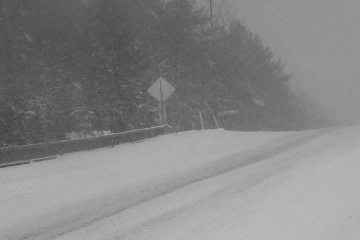}
\includegraphics[width=0.22\linewidth]{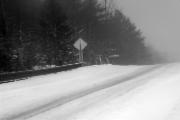}
\includegraphics[width=0.22\linewidth]{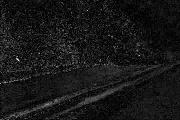}
\includegraphics[width=0.22\linewidth]{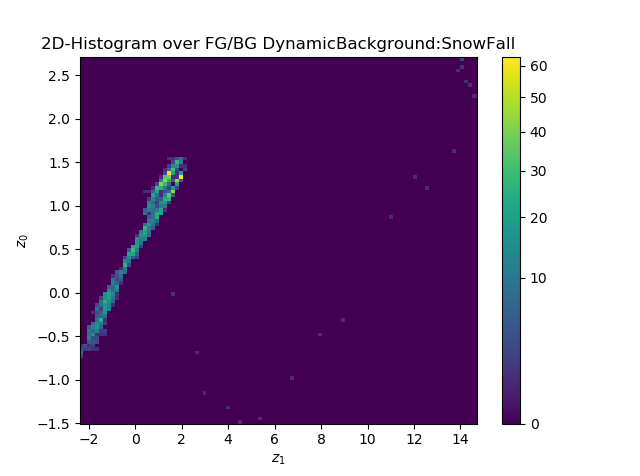}\\
\includegraphics[width=0.22\linewidth]{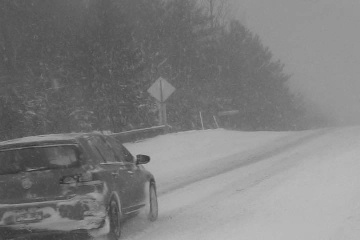}
\includegraphics[width=0.22\linewidth]{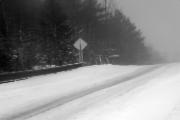}
\includegraphics[width=0.22\linewidth]{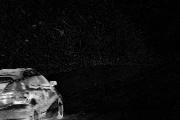}
\includegraphics[width=0.22\linewidth]{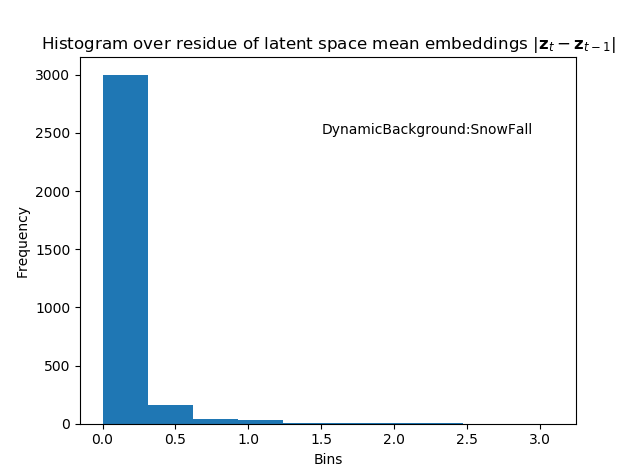}
\end{center}
\vspace{-0.6cm}
  \caption{The input-output pairs and absolute value of residue between input-output pairs from a Convolutional VAE : top half without foreground bottom half with foreground. We remark that the dynamic background such as the snow has been removed. The right column demonstrates the 2d-Histogram over the latent space $\mathbf{z}$ of the CVAE (top) and the histogram over the temporal residue over $\mathbf{z}$ for the same test sequence. }
\label{fig:long}
\end{figure}

VAEs are deep generative models that approximate the distribution for high-dimensional vectors $\mathbf{x}$ that
correspond to pixel values in the image domain. Like a classical auto-encoder. VAEs consists of a probabilistic
encoder $q_\phi(\mathbf{x}|\mathbf{z})$ that reduces the input image to latent space vector $\mathbf{z}$ and
enforces a Gaussian prior, and a probabilisic decoder $p_\theta(\mathbf{x}|\mathbf{z})$ that reconstructs these
latent vectors back to the original images. The loss function constitutes of the KL-Divergence regularization
term, and the expected negative reconstruction error with an additional KL-divergence term between the latent
space vector and the representation with a mean vector and a standard deviation vector, that optimizes the 
variational lower bound on the marginal log-likelihood of each observation \cite{kiran2018overview}.  The classical cascade : CHP, ordered sequence of modes of GMM ($\mu_i, \sigma_i)$, can now be envisaged in the
latent space for a multivariate 1-Gaussian $\mathcal{N}(\mathbf{z}, \mathbf{0}, I)$. The future goal would be to
create Early rejection classifiers as in \cite{zhang2017detecting} for classification tasks, where within each
layer of the probabilistic encoder we are capable of measuring the log-likelihood of being foreground. Storing
previous latent space vectors for the CHP test would require addition memory aside that assigned to the latent
space mean and variance vectors. VAEs are an ideal extension to the rejection cascade since the pixel-level tests in CoG are now performed by the VAE in the latent space, over which a likelihood can be evaluated. We also gain the invariance to positions, orientations, pixel level perturbations, and deformations in mid-level features due the convolutional architecture.
A convolutional VAE with latent space of 16 dimensions was trained on the CDW-2014 datasets \cite{wang2014cdnet}, preliminary results are show in figure \ref{fig:long}.
\vspace{-0.25cm}
\section{Conclusion}
The CoG was evaluated on the wallflower dataset, as well as its autoencoder counterpart VAE-CoG on the CDW-2014 datasets. We observed a speedup of 4-5x, over the baseline GMM, with an average improvement of 17\% in the mis-classification rate.
This study has demonstrated conceptually how a GMM can be re-factored optimally into a prior scene based pixel density and rejection cascade constituent of simpler models ordered based on the probability of occurrences of each level of the cascade, the
accuracy (and complexity) of each model in the cascade level. 

\bibliographystyle{splncs04}
\bibliography{example}
\end{document}